\documentclass[11pt]{article}
\usepackage{amsthm, amsmath, amssymb, amsfonts, url, booktabs, tikz, setspace, fancyhdr, amsbsy}
\usepackage[margin = 0.8in]{geometry}
\usepackage{hyperref, enumerate}
\usepackage{esint}
\usepackage{verbatim}

\usepackage{array}
\usepackage{booktabs}
\usepackage{multirow}
\usepackage{amsmath}
\usepackage{amsthm}

\usepackage{subfig}

\theoremstyle{definition}

\begin{document}

\title{A novel approach for wafer defect pattern classification based on topological data analysis}

\author{Seungchan Ko\thanks{Department of Mathematics, Sungkyunkwan University, Suwon, Republic of Korea. Email: \tt{ksm0385@gmail.com}} , ~Dowan Koo\thanks{Department of Mathematics, Yonsei University, Seoul, Republic of Korea. Email: \tt{dowan.koo@yonsei.ac.kr}}
}

\date{~}

\maketitle

~\vspace{-1.5cm}

\begin{abstract}

In semiconductor manufacturing, wafer map defect pattern provides critical information for facility maintenance and yield management, so the classification of defect patterns is one of the most important tasks in the manufacturing process. In this paper, we propose a novel way to represent the shape of the defect pattern as a finite-dimensional vector, which will be used as an input for a neural network algorithm for classification. The main idea is to extract the topological features of each pattern by using the theory of persistent homology from topological data analysis (TDA). Through some experiments with a simulated dataset, we show that the proposed method is faster and much more efficient in training with higher accuracy, compared with the method using convolutional neural networks (CNN) which is the most common approach for wafer map defect pattern classification. Moreover, our method outperforms the CNN-based method when the number of training data is not enough and is imbalanced. 
\end{abstract}


\section{Introduction}\label{intro}
A Wafer map provides a graphical representation of defect distribution and the defect pattern contains important information to identify the root causes of process failures and any potential problems. After certain steps of the manufacturing process in lines, metrology facilities inspect to detect abnormalities of dies and create a wafer map based on the detected abnormalities. This wafer map visualization help engineers monitor any unusual defect distribution and take appropriate action for the corresponding problems. For example, once a wafer map is produced with some corresponding root causes, similar defect distributions of wafers may indicate the common problem of a certain step, and this would be useful to tackle the problems. 

In modern semiconductor manufacturing, advances in manufacturing technology have brought a lot of positive aspects such as high product performance, cost reduction, and yield improvement. This high-frequency operation has caused more complicated problems and quality variation and it increases the importance of the wafer map defect classification task. It has been done manually by engineers (see, for example, \cite{LC2013}). This manual approach not only takes a long time and needs high labor costs, but also suffers from a lack of consistency because it heavily depends on the engineer's proficiency. In recent years, the automatic classifier of wafer map defect patterns has gained more attention due to various advantages such as better performance, lower cost, and better consistency. A wide range of machine learning algorithms has been studied to solve this problem. See \cite{A2015, YL2016, FQ2016, P2018, vote2019}, where the applications of regression analysis, decision tree, support vector machine, and artificial neural network are addressed. 

In particular, a convolutional neural network (CNN) has gained more popularity and becomes one of the most common methods for this task. CNN is an end-to-end model which does not require any domain-specific feature engineering and has achieved tremendous success in image classification. Since a defect pattern on a wafer map can be represented as an image, CNN-based methods have exhibited reasonably good performance in defect map classification tasks. In this direction, see \cite{NK2018, KK2018, YXW2019} and \cite{SAL2020}.  However, the automatic classifiers including CNN have some drawbacks in several aspects.

Firstly, the CNN-based models contain a large number of parameters and are computationally expensive. It takes a long time to train the model and to predict the defect patterns for a large number of wafers. In the real manufacturing process, however, it is expected to make prompt predictions for a huge number of wafers in a limited time. Furthermore, it is also necessary to retrain the model regularly in accordance with the changes in a manufacturing process or new data accumulation. Hence, it is important to develop a fast and computationally efficient model with a small number of parameters.

Secondly, the performance of the aforementioned classifiers highly depends on the amount and quality of the training data because they are based on the supervised learning algorithm. To train CNN-based models, for example, a large amount of training data is needed. Since the training data of wafer maps can only be obtained through manual inspection, it usually requires high labor costs and takes a long time to prepare the training data. In addition, this labeling task depends on the proficiency of the engineers so the labeling may not be coherent and the quality of the training data may deteriorate. Hence, from a practical point of view, it is desirable to develop a wafer map classifier that can be effectively trained with relatively few training data. 

Thirdly, the training data might not be balanced. In the actual manufacturing process, some defective patterns appear frequently while some defective patterns are rarely found. This problem is worse in the case of well-established processes. It is ideal for the number of wafers for each pattern to be similar, but it is difficult to be satisfied in practice since there are various patterns and each defective pattern stems from different causes in the manufacturing process. To handle this problem, some image augmentation techniques were used for the CNN-based model, but we cannot expect that these augmented data properly reflect the nature of the original defect distributions.

Recently, the area of computational topology has grown rapidly and provides plenty of computational tools for data analysis (\cite{com_top}). One of the common tools is {\textit{persistent homology}} from topological data analysis, which observes the dynamics of topological features in a sequence of nested topological spaces (\cite{ph_1}). This information can be visualized as a persistence diagram (PD), a set of points in the two-dimensional plane where a new topological feature appears at $x$-coordinate and disappears at $y$-coordinate. This compact description of topological features is more useful for the analysis of complex and high-dimensional data because many important topological features are invariant under dimensionality reduction in some sense. It is noteworthy that the spaces of PDs can be equipped with a metric function and it is known that these functions are stable with respect to small perturbations of given data. See, for example \cite{dist_1, dist_2, dist_3, dist_4, dist_5} where {\textit{bottleneck}} and {\textit{Wasserstein}} metrics were adopted and the stability issue was addressed. 

However, computing the bottleneck or Wasserstein distance between PDs is computationally expensive when the number of off-diagonal points is large. Furthermore, to utilize many useful machine learning techniques with PDs, some extra structure other than a metric function is needed. To overcome this problem, there has been plenty of efforts to transform PDs into suitable forms for various machine learning algorithms. One of the most common approaches is a persistence landscape (PL). In \cite{PL}, the author developed a concept of PL, a functional representation $\lambda:\mathbb{N}\times\mathbb{R}\rightarrow[-\infty,\infty]$ of a PD in a Banach space. For $1\leq p\leq\infty$, we define the $p$-landscape distance as $\|\lambda_1-\lambda_2\|_p$ for two PLs $\lambda_1$ and $\lambda_2$. Then the $p$ and $\infty$-landscape metrics are stable with respect to the $p$-Wasserstein distance and to the bottleneck distance respectively on PDs. 

On the other hand, in \cite{PI}, the notion of persistence image (PI) was introduced. PI is a representation of a PD as a finite-dimensional vector and it is proved that this transformation equips the stability as well. PL is sometimes more useful because the mapping from a PD to a PL is invertible. However, PI is more advantageous in many applications because it can be applied with a wide range of machine learning techniques. It was shown in \cite{PI} that PI is computationally more efficient to compute and PIs outperform PLs in various classification and clustering tasks.

Therefore, in this paper, we adopt the PI approach for the representation of PD. Once we extract a topological feature from the given defect pattern on a wafer and represent it as a PI, then we shall use these PIs as inputs of neural networks for classification. As it will be shown later, this approach is computationally efficient, and robust against the small-data and imbalanced-data problems mentioned above, compared with the CNN-based approach.

To the best of our knowledge, this is the first study that applies the techniques from topological data analysis to the wafer map classification problem. Although some shortcomings need to be addressed and further research should be followed-up, this new method has shown better performance in various aspects compared to the existing methods. We expect this TDA-based method to be utilized as a new alternative for the defect pattern classification problem, especially when the training data is insufficient and imbalanced.

The rest of the article is organized as follows. In Section \ref{theory}, we review the theory of persistent homology. We describe how to extract topological structure from the given point-set data and how to represent it as a PD. Moreover, the converting procedure from PD into PI is introduced and the stability of the transformation is discussed. In Section \ref{method}, we shall present in detail our method of expressing wafer defect pattern as a finite-dimensional vector, based on the theory discussed in Section \ref{theory}. In Section \ref{experiment}, some experiments on the classification of wafer map defect patterns will be performed with simulated data and confirm the advantages of the proposed method over the previous methods. Finally, some concluding remarks will be provided in Section \ref{conclusion}.

\section{Persistent homology and vector representation}\label{theory}

We briefly review some essential mathematical theory of homology for completeness. Homology can be used to distinguish different (non-homeomorphic) topological spaces. In particular, simplicial homology can be computed by considering simplicial complex. A simplicial complex $S$ consists of $k$-simplicies, where $k\in\mathbb{N}$. A $0$-(dimensional) simplex is a vertex, $1$-simplex is an edge, $2$-simplex is a triangle, and $3$-simplex is a tetrahedron. In general, $k$-simplex contains $k+1$ vertices and is denoted by $[v_0,v_1,...,v_k]$. The bracket notation implies that the orientation is under consideration. We call a family of simplices $S$ a Simplicial complex if it satisfies the following properties:  
\begin{itemize}
\item for any simplex $\sigma$ in $S$, all lower-dimensional simplices of $\sigma$ are contained in $S$, and
\item the non-empty intersection of any two simplices in $S$ is again a simplex in $S$.
\end{itemize}
For instance, the intersection of two adjacent triangles sharing two vertices ($2$-simplex) is an edge ($1$-simplex). 

Now we consider the vector space $C_k$ which consists of all $\mathbb{R}$-combinations of $k$-simplices of $S$. The boundary map $\partial_k$ acts on a $k$-simplex to its boundary as a sum of its $(k-1)$-dimensional faces, formally defined as 
\begin{equation*}
\partial_k[v_0,v_1,...,v_k]:=\sum_{i=0}^k (-1)^i[v_0,...,\hat{v}_i,...,v_k]
\end{equation*}
where $[v_0,...,\hat{v}_i,...,v_k]$ is a $(k-1)$-simplex obtained from $[v_0,...,v_k]$ with $v_i$ removed. It is noteworthy that the boundary of a boundary is always zero, i.e $\partial_{k-1}\circ\partial_k =0$.

We denote $B_k$ by the image of $\partial_{k+1}$, whose elements are called $k$-boundaries and $Z_k$ by the kernel of $\partial_k$, i.e elements of $C_k$ whose image under $\partial_k$ is zero in $C_{k-1}$. The elements of $Z_k$ are called $k$-cycles. Clearly, $Z_k$ contains $B_k$ because $\partial_k\circ\partial_{k+1} = 0$. And then we define the $k$-homology group of $S$ by $H_k(S):=Z_k/B_k$ and the rank of this group is called the \textit{Betti number}. Typical elements of the homology group are connected components, loops, and spheres. For more details and concrete examples, see \cite{hatcher}.

In topological data analysis, the idea of homology is often employed to classify point-cloud data. In this paper, we shall use \textit{Vietories-Rips} simplicial complexes \cite{ph_1,com_top, intro_tda}. Let $Y$ be a given point-set data equipped with a metric structure so that the distance between any two points in $Y$ is assigned. For a fixed scale $\epsilon>0$, we shall say $k+1$ points in $Y$ form a $k$-simplex if their mutual distance is at most $\epsilon$. We denote $S_{\epsilon}$ by the simplicial complex obtained by this procedure. Though $Y$ is a set of vertices, $S_{\epsilon}$ may contain loops or some interesting homological objects for certain $\epsilon>0$. However, it is not possible to choose such a `nice' $\epsilon>0$ \textit{a priori}. For example, for small $\epsilon>0$, the \textit{Vietories-Rips} complexes would appear as islands of many connected components whereas, for large $\epsilon>0$, the complex would be realized as a single connected component.

To circumvent this difficulty, the idea of persistent homology was introduced \cite{hatcher, com_top, ph_1}. The core idea of persistent homology is to observe the `birth' and `death' of each homological feature along with gradually increasing scales. We denote $S_{\epsilon_1},S_{\epsilon_2},\,\cdots, S_{\epsilon_n}$ by Vietories-Rips complexes with increasing scales $\epsilon_1\leq\epsilon_2\leq\cdots\leq\epsilon_n$. To extract the overall topological feature of a point-set $Y$, we will track the elements of $H_k(S_{\epsilon_i})$ as the scale $\epsilon_i$ augments. 

A persistent diagram (PD) is often used as a standard way to represent the persistent homology as a multi-set of points in $\mathbb{R}^2$. For a fixed dimension $k$, each point $(b,d)\in\mathbb{R}^2$ is encoded as birth and death, where a topological homology appears at scale $b$ and disappears at scale $d$. It is obvious that every point in PD is above the graph $y=x$ since every `death' happens after `birth'. Usually, the points far from the diagonal line are considered to be a robust topological property as the length $d-b$ denotes its persistence. It is noteworthy that the space of persistence diagrams can be equipped with some metrics. One of the most common metrics used for PDs is the $p$-{\textit{Wasserstein distance}} between two PDs $B$ and $B'$, which is defined by
\begin{equation*}
W_p(B,B')=\inf_{\gamma: B \to B'} (\sum_{u\in B} \|u-\gamma(u)\|_{\infty}^p)^{1/p}
\end{equation*}
where $1\leq p <+\infty$. When $p=\infty$, then $W_{\infty}$ metric is called {\textit{bottleneck distance}} and the $\sum_{u\in B}$ is replaced by $\sup_{u\in B}$. These metrics can be used to compare the (dis-)similarity of two PDs and will be utilized to represent the stability of the vector transformation later. See, for example, \cite{dist_intro_1, dist_intro_2} where a general study of the space of persistence diagrams
endowed with $W_p$ metrics was studied.

As mentioned earlier, we shall use Persistence Image (PI) as a converting method from a PD into a finite-dimensional vector as proposed in \cite{PI}. Let $B$ be a given PD in birth-death coordinates. We transform $B$ via a map $T$ defined by $T(x,y):=(x,y-x)$. Hence, the image $T(B)$ can be interpreted as a birth-persistence diagram. Next, let $\phi_u$ be a probability distribution with mean $u=(u_x,u_y)\in \mathbb{R}^2$. In this paper, we will choose $\phi_u$ to be the Gaussian distribution with mean $u\in T(B)$ and variance $\sigma^2$. Similarly, as before, the points near the horizontal axis are considered to be noise whereas the points away from the horizontal axis are rather robust. This motivates us to consider the use of weighting function $f:\mathbb{R}^2\to \mathbb{R}$, which is zero along the horizontal line, continuous and piecewise differentiable. For instance, one common option is to take $f$ as
\begin{equation}\label{wt_fn}
f(u_x,u_y):=\begin{cases}
1\quad&{for}\quad u_y >c\\
\frac{u_y}{c} \quad&{for}\quad 0\leq u_y\leq c\\
0\quad&{for}\quad u_y<0
\end{cases}
\end{equation}
For the converted PD by the mapping $T$ together with the weighting function $f$, we now define the persistent surface by
\begin{equation}\label{surface}
\rho_B(z):=\sum_{u\in B}f(u)\phi_u(z)
\end{equation}
Here the choice of the weighting function is crucial in proving the stability from PDs to PIs. 

Now, as a final step, we discretize the persistent surface $\rho_B$ by taking averages over the boxes. We first fix any grid with $n$ boxes, where each box(or pixel) is denoted by $p$. We then define the persistent image by $I(\rho_B)_p=\int_p \rho_B \,{\rm{d}}x{\rm{d}}y$. For a given PD $B$, we define its persistent image by the collection of pixels $I(\rho_B)_p$, and therefore we have obtained the finite-dimensional vector representation of the given PD.
 
Lastly, we discuss the stability of the mapping described above. Here, stability means that the distance between two output vectors can be controlled by the perturbation of corresponding input data. In practice, a lack of stability might lead to a tragic situation when a small change in input can make violent behavior for the output, which significantly lowers the validity of the given transformation.

Fortunately, the stability estimate is available for the mapping from PDs to PIs where the input data PDs are equipped with $1$-Wasserstein distance. Since PIs are vectors, it is natural to assign the Euclidean distance. More precisely, if we use the Gaussian distribution in \eqref{surface}, we can deduce the following stability estimate which is quoted from \cite{PI}:
\begin{equation}\label{stability_PI}
\|I(\rho_B)_p-I(\rho_{B'})_p\|_1\leq (\sqrt{5}\|\nabla f\|_{\infty}+\sqrt{\frac{10}{\pi}}\frac{\|f\|_{\infty}}{\sigma})W_1(B,B'),
\end{equation}
where $f$ is the weight function and $\sigma$ is the standard deviation of the Gaussian distribution. In the next subsequent section, we will describe the way in detail to transform a given wafer map into the PI.

\section{Methodology}\label{method}

\begin{figure}[h]
\centering
\includegraphics[width=1\textwidth ]{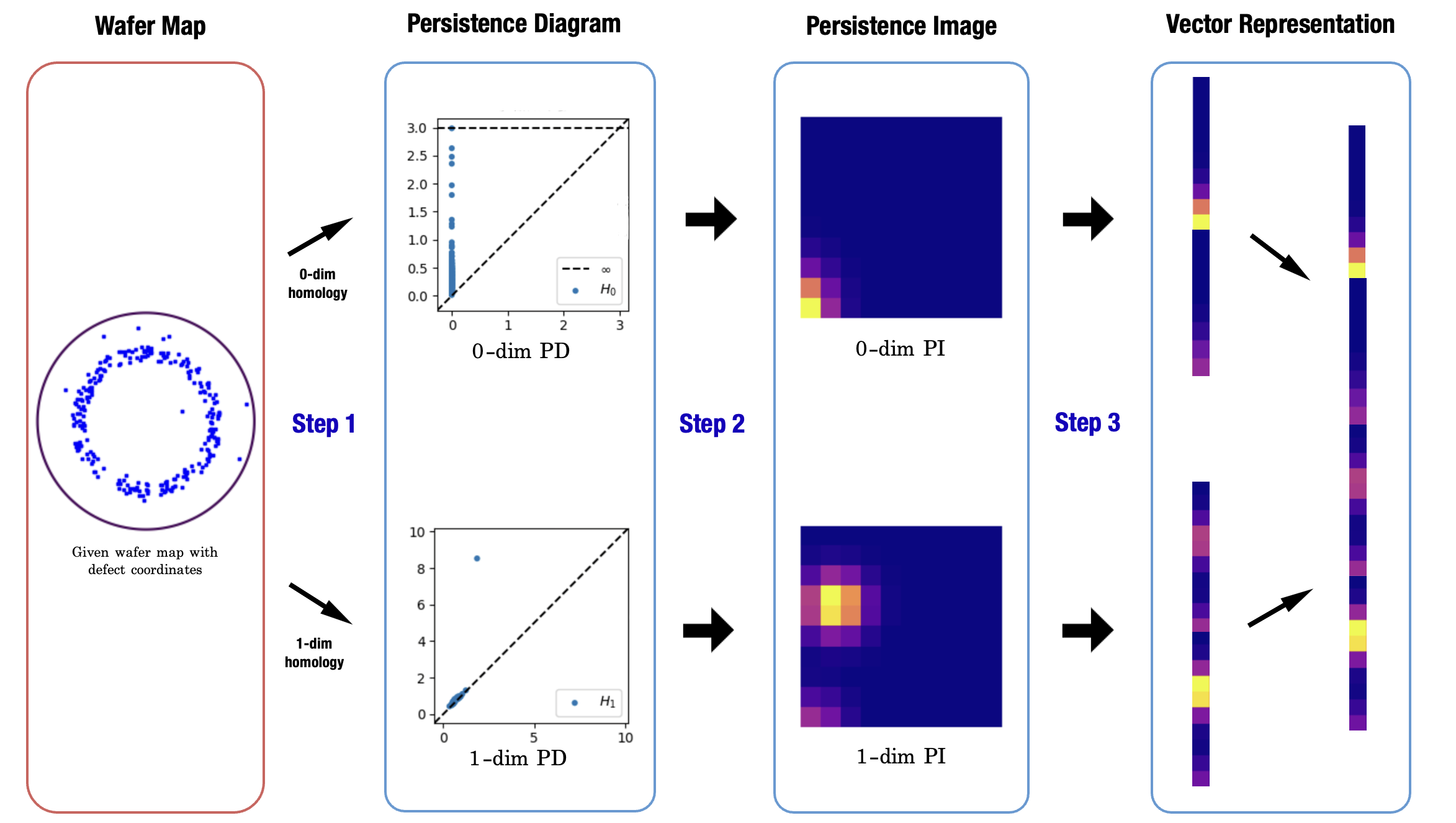}
\caption{ 
The entire process of our data preprocessing method based on topological data analysis. The `Step'  in the figure corresponds to the `Step' listed in Section \ref{method}.
\label{overall_fig}}
\end{figure}

In this section, we present our main methodology to obtain a finite-dimensional vector from a given wafer map based on the theory described in Section \ref{theory}. This vector encodes the overall topological features of the wafer map. Our data preprocessing method consists of the following three steps; from the wafer map, the zero and one-dimensional persistence diagrams (PDs) are obtained. Secondly, these PDs are converted into persistence images (PIs) for each dimension. Lastly, after flattening those two PIs to vectors, we shall concatenate them into a single vector of finite length. The entire process is depicted in Figure \ref{overall_fig}. 

\begin{itemize}

\item[Step 1] - Obtaining Persistence Diagrams

\hspace{5mm}Firstly, we shall convert the given wafer map into a PD. Since every wafer map itself is two-dimensional, we only need to collect the zero and one-dimensional PDs because higher-dimensional PDs are trivial. Note that a wafer map can be interpreted as a collection of dots on a disk and by endowing radii to these dots, they can be connected to the neighbors and gradually conglomerate, as the scale increases. Each point in PD represents the scale at which the corresponding topological feature (e.g., connected component, loop) appears and disappears. The PD can exhibit the topological properties of wafer map defect patterns. The zero-dimensional PD can shed light on the local/global closeness of the dots, while one-dimensional PD can effectively capture the loop patterns formed by the dots. For this process, we use the Python package Ripser \cite{ripser_1, ripser_2, scklearn_tda}.

\hspace{5mm}In $0$-dim PD in Figure \ref{overall_fig}, all the points align vertically with birth scale $0$. This is because, in the beginning, every dot in the wafer forms a connected component and some dots converge into the same component as the radius increases, which results in the disappearance of the zero-dimensional topological feature (connected component). The point $(0,3.0)$ designates that every dot in the wafer has at least one another dot within the distance at most $3$. More importantly, the density of points in $0$-dim PD from Figure \ref{overall_fig} is relatively high in the lower part of the diagram; it means that most of the dots have other dots nearby. 

\hspace{5mm}In $1$-dim PD from Figure \ref{overall_fig}, a point around $(2,9)$ in the north-western corner stands out while other points are clustered near the origin. The points near the $y=x$ would be considered less relevant but the points far away from this diagonal line are more crucial in unveiling the essential structure of the wafer map. The point near $(2,9)$ indicates that some collection of points start to form a loop at around radius $2$ and lose that shape at around radius $9$. The far off-diagonal points in $1$-dim PD are often related to loops or curvy lines of wafer maps.

\item[Step 2] - Transforming Persistence Diagrams into Persistence Images

\hspace{5mm}In this stage, PDs are converted into PIs. Although this procedure was mentioned in Section \ref{theory}, we briefly discuss again this process with precise parameters used in the experiments which will be conducted in Section \ref{experiment}. First, PD is transformed into a persistent surface. More precisely, each point $(u_x,u_y)$ in the PD is shifted to the birth-persistence coordinate by $T(u_x,u_y)$, where $T(x,y)=(x,y-x)$. Here, $y-x$ can be interpreted as the persistence of its topological feature. Then, we consider the Gaussian probability distribution of mean $T(u_x,u_y)$ and variance $\sigma^2$. The variance is taken as $\sigma^2= 0.01$ in this paper. Summing all the Gaussian functions corresponding to each shifted point of  PD multiplied by the linear weighting function, we get the persistent surface. As the persistent surface is continuous, and hence the resolution is very high. Therefore we need to discretize the surface to generate the finite-dimensional vector from it. The entire domain of the persistent surface is partitioned into many pixels of the same size. The number of pixels is taken to be $20\times20$ in the experiments in Section \ref{experiment}. By taking an average of the persistent surface over each pixel, we get the persistent image (PI). In the PI depicted in Figure \ref{overall_fig}, the bright yellow and reddish part designates high values while the dark blue part indicates lower values. For this transformation of PD into PI, we will use the Python library Persim \cite{scklearn_tda}.

\item[Step 3] Vector representation of PIs

\hspace{5mm}At the final preprocessing step, we shall rearrange the components of PI to make a finite-dimensional vector. Each zero and one-dimensional PI encoded in matrix forms (as each PI consists of $20\times20$ pixels) are flattened as two $400$-dimensional vectors. And then these two vectors are concatenated into a single $800$-dimensional vector. This final output vector will be used as the input data of the neural network for the classification task. 
\end{itemize}

We have described our method to transform a given wafer map into a finite-dimensional vector. One striking difference between this representation methodology and the CNN-based feature extraction is the dimension of resulting vectors. As mentioned earlier, our TDA-based method generates an $800$-dimensional vector, while the CNN-based method using ResNet50 generates a vector of $51200$ components. Surprisingly, as we will see in Section \ref{experiment}, this lower-dimensional vector generated by the proposed method encapsulates the information of the defect distribution very well, and the classification model with these vectors performs much better than the model with vectors generated by CNN.

\section{Experiments}\label{experiment}
In this section, we will conduct several experiments to demonstrate the performance of our proposed method. For the experiments, we shall use simulated datasets. More precisely, we artificially generate the defect patterns to train the classifier and evaluate the performance of the proposed method. We have selected five types of most common defect patterns by referring to the experiments performed in other research papers \cite{NK2018, KK2018, YXW2019}, and by consulting some domain experts or open-source data, e.g. WM-811K wafer dataset, which is the largest open-source dataset from the real manufacturing process: random pattern and non-random patterns (ring, scratch, dense and cluster). From the yield analysis point of view, wafer maps with non-random distributions are important and should be carefully inspected, because they indicate the problematic issues related to the manufacturing process. On the other hand, random patterns with a controllable number of defects are acceptable, and they only provide less amount of information from the perspective of process improvement.

\subsection{Data generation}\label{data}
\begin{figure}
\centering
\subfloat[Random pattern]{{\includegraphics[width=1\textwidth ]{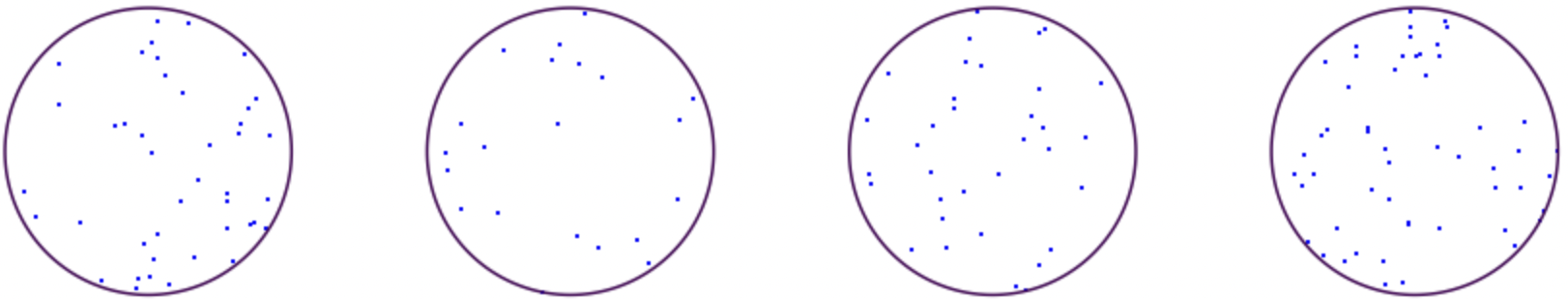} }}%
\hfill 
\centering
\subfloat[Ring pattern]{{\includegraphics[width=1\textwidth ]{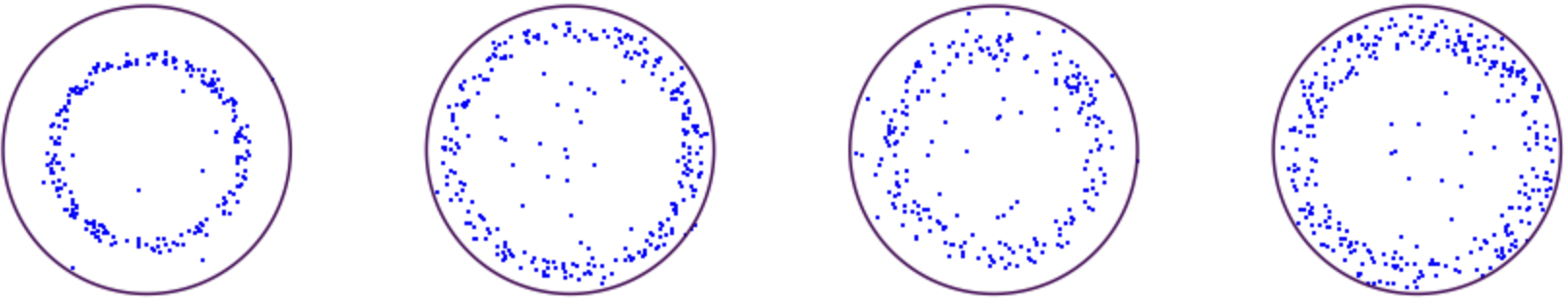} }}%
\hfill 
\centering
\subfloat[Scratch pattern]{{\includegraphics[width=1\textwidth ]{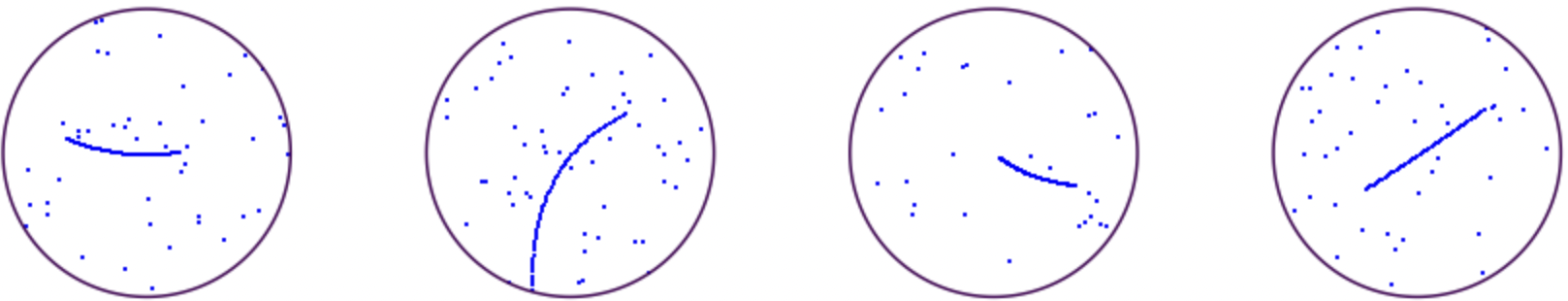} }}%
\hfill 
\centering
\subfloat[Dense pattern]{{\includegraphics[width=1\textwidth ]{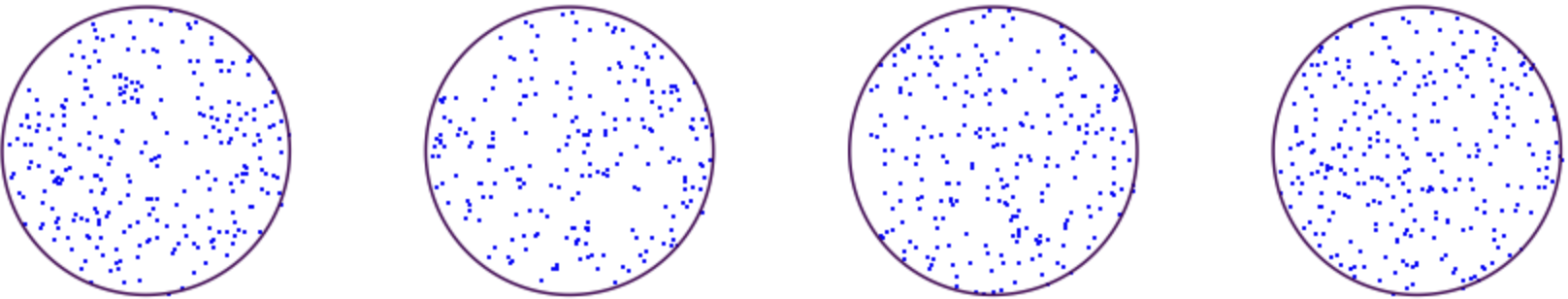} }}%
\hfill 
\centering
\subfloat[Cluster pattern]{{\includegraphics[width=1\textwidth ]{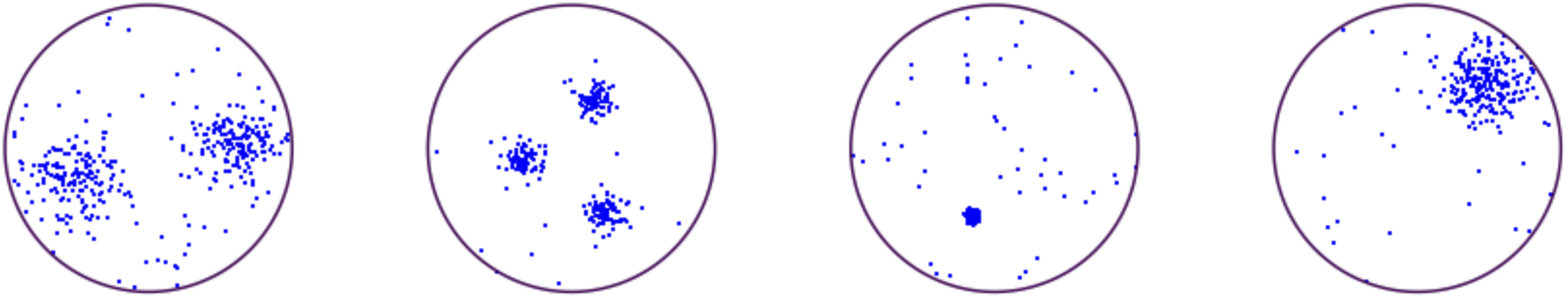} }}%
\caption{Some samples of datasets generated as outlined in Subsection \ref{data} for (a) random pattern, (b) ring pattern, (c) scratch pattern, (d) dense pattern, (e) cluster pattern.}%
\label{sample}%
\end{figure}

Let us now describe the detailed procedure of the defect-pattern generation. We first discuss the way to generate wafer maps with a random pattern. As mentioned above, this random distribution itself will be used as one data category (random). Moreover, these randomly generated defects will also be used as noise in generating non-random datasets (ring, scratch, dense, and cluster). A wafer will be displayed in $\mathbb{R}^2$ by a disk with center $(0,0)$ and radius $10$.
\begin{itemize}
\item Random: For the random pattern category, we randomly choose the number of defects for each wafer by $n_{\rm{random}}\sim{\rm{Uniform}}(10,60).$ We shall then use the polar coordinate to represent each defect on a wafer map. The angle and the radius are generated as
\[\theta_{\rm{random}} \sim {\rm{Uniform}}(0,2\pi)\quad{\rm{and}}\quad r_{\rm{random}}\sim {\rm{Uniform}}(0,10).\]
\item Ring: We again use the polar coordinate system to represent the defects of the ring pattern. The number of ring-pattern defects is randomly chosen by $n_{\rm{ring}}\sim{\rm{Uniform}}(150,300)$. Then we generate the pattern as
\[\theta_{\rm{ring}} \sim {\rm{Uniform}}(0,2\pi)\quad{\rm{and}}\quad r_{\rm{ring}}\sim {\rm{Uniform}}(r_0,r_0+\delta),\]
where the inner radius $r_0>0$ and width $\delta>0$ of a ring are randomly given by
\[r_0 \sim {\rm{Uniform}}(3,6)\quad{\rm{and}}\quad \delta\sim {\rm{Uniform}}(0,4).\]

\item Scratch: We shall generate this pattern based on the quadratic function
\begin{equation}\label{curve}
y=kx^2\qquad{\rm{on}}\,\,\,[a,b],
\end{equation}
where the parameters are randomly chosen by
\begin{align*}
a,\,b&\sim{\rm{Uniform}}(-10,10)\quad{\rm{and}}\quad|a-b|>5,\\
k&\sim{\rm{Uniform}}(-1/15,1/15)\quad{\rm{and}}\quad k\neq0.
\end{align*}
We choose $n_{\rm{scratch}}\sim{\rm{Uniform}}(50,100)$, and select $n_{\rm{scratch}}$ evenly distributed points $\{x_i\}^{n_{\rm{scratch}}}_{i=1}$ in $[a,b]$.
If we let $y_i=kx_i^2$ for all $i\in\{1,2,\cdots,n_{\rm{scratch}}\}$, we obtain the set of points $\{(x_i,y_i)\}^{n_{\rm{scratch}}}_{i=1}$ lying
on the quadratic curve \eqref{curve}.
Finally, we obtain the points $\{(x'_i,y'_i)\}^{n_{\rm{scratch}}}_{i=1}$ describing the random scratch by randomly rotating the portion of curve represented as $\{(x_i,y_i)\}^{n_{\rm{scratch}}}_{i=1}$:
\[
\begin{bmatrix}
x'_i\\
y'_i\\
\end{bmatrix}
=
\begin{bmatrix}
\cos\theta & -\sin\theta\\
\sin\theta & \cos\theta\\
\end{bmatrix}
\begin{bmatrix}
x_i\\
y_i\\
\end{bmatrix}
\quad\forall i\in\{1,2,\cdots,n_{\rm{scratch}}\},
\]
where 
\[\theta\sim{\rm{Uniform}}(0,2\pi).\]

\item Dense: We generate this pattern of defects in the same way as we did for the random pattern generation, but the defects are more densely distributed. We randomly select the number of defects by $n_{\rm{dense}}\sim{\rm{Uniform}}(150,300)$, and the defects are represented by the polar coordinate 
\[\theta_{\rm{dense}} \sim {\rm{Uniform}}(0,2\pi)\quad{\rm{and}}\quad r_{\rm{dense}}\sim {\rm{Uniform}}(0,10).\]

\item Cluster: In order to generate defects in this category, we shall use the function {\textit{make-blobs}} from the scikit-learn library \cite{scklearn}. The number of sample is chosen by $n_{\rm{cluster}}\sim{\rm{Uniform}}(150,300)$, the number of clusters is randomly determined among $\{1,2,3\}$ with the probability $\frac{1}{3}$ for each, and the standard deviation of the clusters follows the distribution ${\rm{Uniform}}(0.1,2)$. 
\end{itemize}
As mentioned earlier, the random noise $n_{\rm{random}}\sim{\rm{Uniform}}(10,60)$ is concatenated to the wafer maps with non-random defect patterns. Some samples of wafer maps for each class generated by the way above are depicted in Figure \ref{sample}.

To evaluate the performance of the proposed method, we shall perform various experiments with simulated data, and compare the results from the TDA-based method and CNN-based method. As described in Section \ref{method}, the given wafer map is transformed into an $800$-dimensional vector. We shall use this vector as an input for the neural network with a single hidden layer with 1024 neurons, and apply the {\textit{ReLU}} activation function. And then {\textit{Adam }}optimizer which is a stochastic gradient-based optimizer was used for parameter updates. For the CNN-based method, we will use the pre-trained ResNet50 (trained on ImageNet) with the single layer of 1024 nodes on the top \cite{resnet}. For this purpose, we first convert the given wafer map into an image and put the image into the ResNet50 model which generates $51200$-dimensional features. Then these features are utilized as input for the layer on the top, which will be trained with the Adam optimizer.

\subsection{Basic performance evaluation}
We will first conduct a basic performance evaluation. For this purpose, 500 wafer maps are generated for each class as described in Subsection \ref{data}. We then split these data into 300 training dataset, 100 validation dataset, and 100 test dataset. The 300 defect patterns for each category are used for training both our proposed model and the CNN model, and the 100 data for each class are used for the validation. Once high enough training/validation accuracies are achieved ($700$ epochs in this paper), the 100 test wafer maps for each category are used to evaluate both models. As discussed in Section \ref{method}, the proposed method in this paper has transformed the given wafer map into an $800$-dimensional vector, while the CNN architecture (ResNet50) has changed the image of the defect pattern to a $51200$-dimensional vector.

\begin{figure}
\subfloat[TDA-based model]{{\includegraphics[width=0.5\textwidth ]{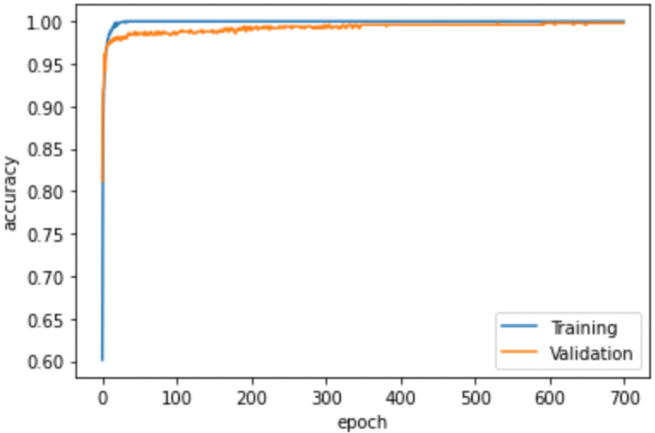} }}%
\subfloat[CNN-based model]{{\includegraphics[width=0.5\textwidth ]{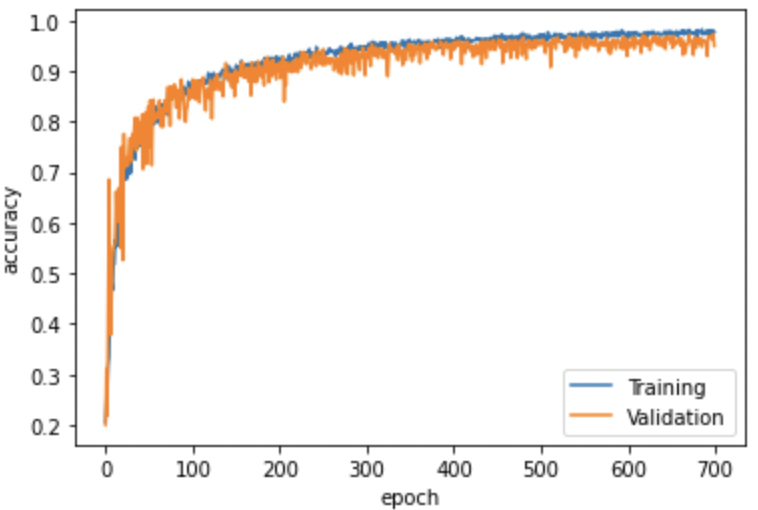} }}%
\caption{The training and validation accuracy for (a) TDA-based method and (b) CNN-based method. Both models are trained with 300 training data for each class and the validation accuracy is estimated with 100 validation data for each category.}%
\label{train_val}%
\end{figure}

\begin{figure}
\subfloat[TDA-based model]{{\includegraphics[width=0.5\textwidth ]{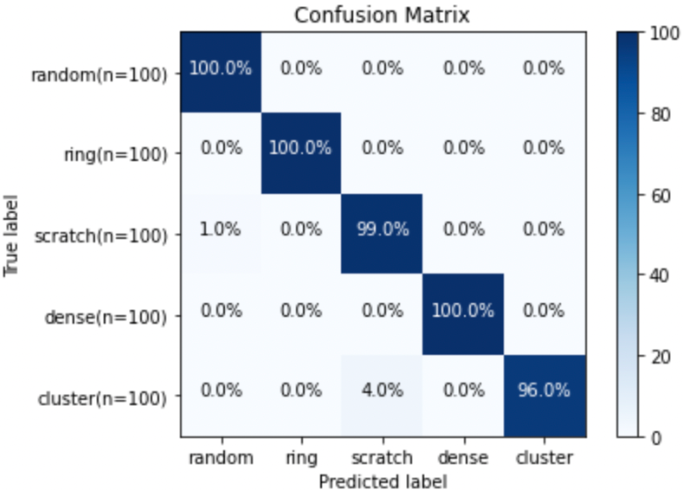} }}%
\subfloat[CNN-based model]{{\includegraphics[width=0.5\textwidth ]{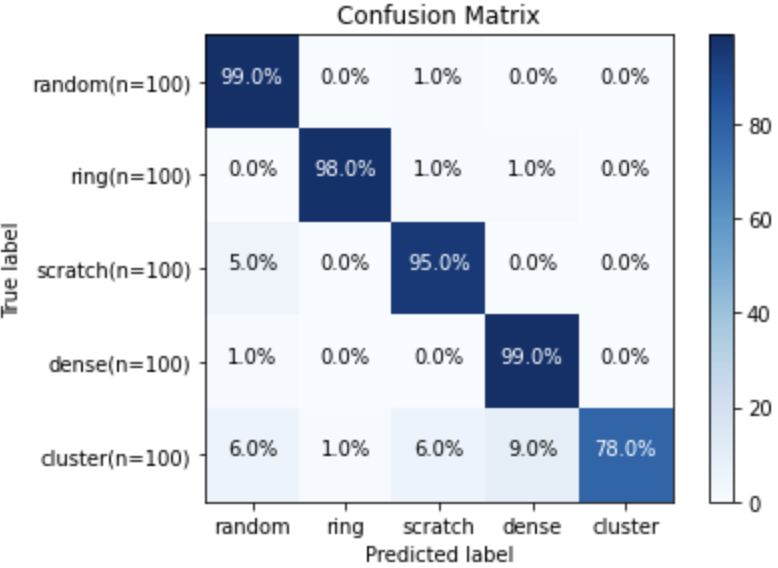} }}%
\caption{Test accuracy confusion matrices for (a) TDA-based method and (b) CNN-based method.}%
\label{confusion}%
\end{figure}

First, the training and validation accuracies for each epoch are shown in Figure \ref{train_val} and the confusion matrices are presented in Figure \ref{confusion} for the test dataset. The overall accuracy for the TDA-based model is $99.0\%$ and for the CNN-based model is $93.8\%$ as presented in Table \ref{basic_1}. Note that although our TDA-based model performs better in accuracy, the difference is modest. However, our method outperforms the CNN-based method in several aspects if we consider the situation in the real manufacturing process.

\begin{table}
\centering
\begin{tabular}{c|ccc}
\noalign{\smallskip}\noalign{\smallskip}\hline\hline
& Accuracy & Training time for each epoch & Epochs needed for $90\%$ accuracy \\
\hline
TDA-based method & $99.0\%$ & $0.012$ sec & $3$ epochs \\
\hline
CNN-based method & $93.8\%$ & $4.234$ sec & 156 epochs \\
\hline
\hline
\end{tabular}
\caption{Basic performance evaluation I: A basic performance comparison was conducted between the TDA-based method and CNN-based method. Both models are trained with 300 training data for each category.\label{basic_1}}
\end{table}

\begin{table}
\centering
\begin{tabular}{c|cc|cc|cc}
\hline\hline
Ratio of random pattern & \multicolumn{2}{c|}{70\%} & \multicolumn{2}{c|}{80\%} & \multicolumn{2}{c}{90\%} \\
\hline
Number of wafer maps   & 500         & 1000        & 500         & 1000        & 500         & 1000       \\ \hline
TDA-based method       & 18.8 sec   & 37.8 sec   & 16.2 sec   & 33.3 sec   & 15.7 sec   & 30.7 sec  \\ \cline{1-1}
CNN-based method       & 35.7 sec   & 67.9 sec   & 36.9 sec   & 69.2 sec   & 35.1 sec   & 67.6 sec  \\ 
\hline\hline
\end{tabular}
\caption{Basic performance evaluation II: The prediction times when the ratio of random defect patterns are 70\%, 80\%, and 90\% are measured. For each portion of the random pattern, the experiments are conducted to make predictions for both 500 and 1000 wafer maps.\label{basic_2}}
\end{table}

One of the most important factors that should be considered in practice is how fast the model can make predictions. In the actual manufacturing process, a huge amount of data is poured, and it is necessary to perform real-time classification on a large number of wafer maps simultaneously. The average times taken for the models to produce a prediction result from raw data are summarized in Table \ref{basic_2}. In practice, the ratio of random defect patterns is very high compared with that of other patterns. Hence the datasets used here consist of the relatively high portion of wafer maps with random patterns: $70\%$, $80\%$ and $90\%$ of random patterns, and equal rates for each remaining non-random pattern, where the wafer maps are randomly generated as outlined in Subsection \ref{data}. As can be seen from Table \ref{basic_2}, our model is much faster compared with the CNN-based model. In particular, while the prediction times needed for CNN-based methods slightly vary, the prediction times clearly decrease as the ratio of random pattern increase. Note that if the ratio of the random pattern is more than $80\%$, the TDA-based model is more than twice faster than the CNN-based method, and therefore it is expected that our method will perform better for real-time classification in the actual manufacturing process.

\begin{figure}%
\subfloat[Training accuracy]{{\includegraphics[width=0.5\textwidth ]{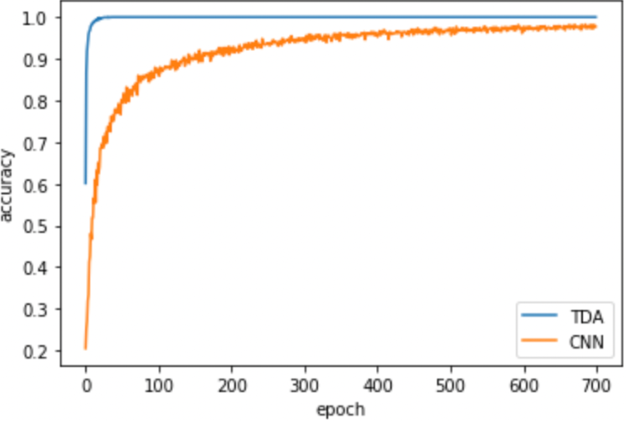} }}%
\subfloat[Training loss]{{\includegraphics[width=0.51\textwidth ]{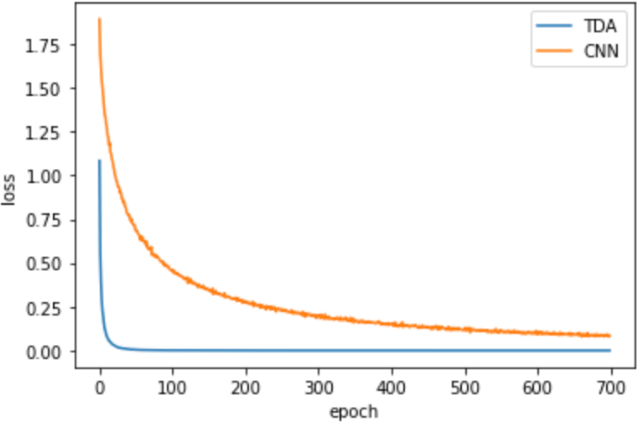} }}%
\caption{A comparison between the TDA-based model and CNN-based model for (a) training accuracy and (b) training loss.}%
\label{comparison}%
\end{figure}

Another important factor is training efficiency. In real manufacturing circumstances, it is often necessary to retrain the model in accordance with the manufacturing situation and the engineers' opinions. For example, one may need to add a new category to the dataset or obtain new training data refined by experienced engineers. Therefore, training efficiency should also be considered in the design of the wafer map classification model. As we can see from Figure \ref{comparison}, we can confirm that our method trains the single-layer perceptron model much faster and more efficiently with fewer epochs. For example, as summarized in Table \ref{basic_1}, the CNN-based method needs at least $156$ epochs to achieve $90\%$ of training accuracy, while our proposed model reaches that accuracy as soon as the training starts. Moreover, the average processing time for each epoch for our model is 0.012 seconds, while the CNN-based model takes 4.234 seconds per each epoch on average.

 As we can see from the above experiments, our method performs better in terms of model accuracy, 
and it also shows overwhelming performance in terms of training time and efficiency. In order to further highlight the better performance of our method, we shall address the issues brought up in Section \ref{intro} and design two more experiments: a small-data experiment and an imbalanced-data experiment, which are covered in the next subsections.

\subsection{Small-data experiment}
In this experiment, we will investigate whether the proposed method can train the neural network sufficiently well only with a small amount of training data. One of the biggest difficulties in the supervised learning problem in practice is obtaining a sufficient amount of quality training data. In particular, a CNN architecture requires a huge amount of training data to train a large number of parameters. In the defect map classification problem, it is difficult to obtain a sufficient amount of data because the training data should be manually labeled by an experienced engineer. In addition, when many engineers simultaneously prepare the training data, the labeling criteria may vary depending on their proficiencies, which may impair the consistency of the data. Therefore, in a wafer map classification task, it would be extremely useful if we can train the classifier only with a small amount of data. Throughout the experiments in this subsection, we will show that the proposed method can train the neural network classifier well enough, even with very few training data. 

For this purpose, we first prepare one dataset and use it in common to evaluate the performance of each model trained with different amounts of training data. This test dataset which consists of 100 wafer maps for each class is generated according to the method described above. Next, we will train the classifier for both TDA and CNN-based methods, with a very small amount of training data at the beginning (10 wafer maps for each) and gradually increase the number of training data (until 100 wafer maps for each). And then we will observe the changes in test accuracies. The result of the experiments is summarized in Table \ref{small_data}.
As the experiment shows, our method can train the classifier well enough with few training data. Furthermore, as the amount of training data decreases, the model trained with the TDA-based method showed only a slight decrease in accuracy ($96.4\%$ to $89.6\%$), whereas the accuracy of the CNN-based model decreased significantly ($90.5\%$ to $66.8\%$).

\begin{table}
\centering
\begin{tabular}{c|cccccccccc}
\noalign{\smallskip}\noalign{\smallskip}\hline\hline
Number of training\\ data (for each class)& $10$ & $20$ & $30$ & $40$ & $50$ & $60$ & $70$ & $80$ & $90$ & $100$\\
\hline
TDA-based model & $89.6\%$ & $91.6\%$ & $93.4\%$ & $95.4\%$ & $97.0\%$ & $96.2\%$ & $95.8\%$ & $96.0\%$ & $97.4\%$ & $96.4\%$\\
\hline
CNN-based model & $66.8\%$ & $74.3\%$ & $71.4\%$ & $73.5\%$ & $79.1\%$ & $81.8\%$ & $85.0\%$ & $86.4\%$ & $87.2\%$ & $90.5\%$\\
\hline
\hline
\end{tabular}
\caption{Test accuracies for both models trained with a small amount of training data}
\label{small_data}
\end{table}

\subsection{Imbalanced-data experiment}
As we mentioned in Section \ref{intro}, in real manufacturing circumstances, it is extremely difficult to obtain a well-balanced dataset. Some wafer map patterns appear frequently, while some other patterns might be found very rarely. This is mainly because each defective pattern on a wafer map stems from different causes in the manufacturing process. Therefore, it is very important to find a way to train a model even with imbalanced data. In this subsection, we will show that our method is much more robust against imbalanced data than the CNN-based method.

For the experiment, $100$ test data for each class of defective pattern are initially generated and fixed throughout the experiments. Next, we shall generate the training data. Unlike previous experiments, we randomly select the number of data in each category. The number of data for each class is randomly chosen by 
\[
N_{(\rm{random})}, N_{(\rm{ring})}, N_{(\rm{scratch})}, N_{(\rm{dense})}, N_{(\rm{cluster})}\sim {\rm{i.i.d.}}\,\, {\rm{Int}}({\rm{Uniform}}(1,300)),
\] 
where, for example, $N_{(\rm{random})}$ denotes the number of data in random pattern category. We iterate this process 10 times so that we have 10 training datasets. We then train the models (both TDA-based and CNN-based) with these different datasets. Finally, we evaluate the accuracy of each model with the initially fixed test dataset. The result of this experiment is presented in Table \ref{undata}.

\begin{table}
\centering
\begin{tabular}{c|ccccc|cc}
\noalign{\smallskip}\noalign{\smallskip}\hline\hline
\multirow{2}{*}{} & \multicolumn{5}{c|}{Number of training data} & \multicolumn{2}{c}{Test accuracy} \\
\cline{2-8}
      & Random  & Ring & Scratch & Dense & Cluster & TDA-based model & CNN-based model \\
\hline
 Dataset 1 & $253$ & $151$ & $102$ & $142$ & $46$ & $95.0\%$ & $81.8\%$ \\
 Dataset 2 & $187$ & $39$ & $210$ & $35$ & $193$ & $97.4\%$ & $84.2\%$ \\
 Dataset 3 & $21$ & $281$ & $49$ & $71$ & $229$ & $90.8\%$ & $66.8\%$ \\
 Dataset 4 & $167$ & $31$ & $212$ & $47$ & $86$ & $96.0\%$ & $81.0\%$ \\
 Dataset 5 & $273$ & $42$ & $5$ & $8$ & $285$ & $81.4\%$ & $57.8\%$ \\
 Dataset 6 & $75$ & $61$ & $29$ & $59$ & $51$ & $90.6\%$ & $83.6\%$ \\
 Dataset 7 & $91$ & $79$ & $90$ & $242$ & $155$ & $96.4\%$ & $87.4\%$ \\
 Dataset 8 & $160$ & $175$ & $253$ & $208$ & $202$ & $98.2\%$ & $94.8\%$ \\
 Dataset 9 & $270$ & $9$ & $43$ & $33$ & $219$ & $93.4\%$ & $72.4\%$ \\
 Dataset 10 & $16$ & $10$ & $96$ & $252$ & $145$ & $86.8\%$ & $54.8\%$ \\
\hline
\hline
\end{tabular}
\caption{Test accuracies for both models trained with imbalanced datasets.}
\label{undata}
\end{table}

As we can see from Table \ref{undata}, we can confirm that our proposed method performed much better than the CNN-based method. When the amount of training data is not very small (Dataset 8), the difference is not significant. However, if training data for certain classes are not enough, our TDA-based method outperforms the existing CNN-based method. In particular, when the number of data in some categories is extremely small (Dataset 3, Dataset 5, Dataset 9, Dataset 10), the CNN-based model does not perform well enough to classify wafer maps normally, whereas our method still achieved high accuracy.

\section{Conclusion}\label{conclusion}
This paper proposes a novel method to classify wafer defect patterns based on the theory of persistent homology from topological data analysis. It is a data preprocessing method to convert a wafer map into a finite-dimensional vector, which is used for the input of neural network-based classification algorithms. The main idea is to impose the topological structure to the given point-cloud data using a simplicial complex and extract the topological properties by observing the dynamics of persistent homology. We then represent the extracted topological features as a finite-dimensional vector and use it as an input for the classifier. Throughout the various experiments with the simulated dataset, we confirm that our method outperforms the most common existing method, namely, the CNN-based image classification approach, in many aspects.

The main contribution of the present paper is that we have developed a new approach for the wafer map classification task, which is completely different from the existing methods. This new methodology is computationally efficient with high model accuracy and robust against the lack of training data and imbalance of training data. This paper serves as a starting point for TDA-based wafer map classification, and we expect that the proposed method has considerable potential to be extended in several directions. One interesting future research direction is refining this preprocessing method to make the resulting vectors work well with the unsupervised clustering algorithms.

\bibliographystyle{abbrv}
\bibliography{ref}

\end{document}